\documentclass[review,11pt]{elsarticle} 
\usepackage{algorithm}
\usepackage{algorithmic}
\usepackage{graphicx}
\usepackage{comment}
\usepackage{amsmath}
\usepackage{subfigure}
\newtheorem{definition}{Definition}
\usepackage{booktabs} 
\usepackage{color}
\usepackage[english]{babel}
\usepackage{picinpar}
\usepackage{moresize}
\usepackage{amssymb}
\usepackage{url}

\DeclareMathOperator*{\argmax}{arg\,max}

\newcommand{\yz}[1]{\textcolor{black}{#1}}

\begin{document}
\begin{frontmatter}

\title{An Online Reinforcement Learning Approach to Quality-Cost-Aware Task Allocation for Multi-Attribute Social Sensing}
\author{Yang Zhang}
\author{Daniel Zhang}
\author{Nathan Vance}
\author{Dong Wang}
\address{Department of Computer Science and Engineering\\
University of Notre Dame\\
Notre Dame, IN 46556}
\begin{abstract}
Social sensing has emerged as a new sensing paradigm where humans (or devices on their behalf)  collectively report measurements about the physical world. This paper focuses on a quality-cost-aware task allocation problem in multi-attribute social sensing applications. The goal is to identify a task allocation strategy (i.e., decide when and where to collect sensing data) to achieve an optimized tradeoff between the data quality and the sensing cost. 
While recent progress has been made to tackle similar problems, three important challenges have not been well addressed: (i) ``online task allocation": the task allocation schemes need to respond quickly to the potentially large dynamics of the measured variables in social sensing; (ii) ``multi-attribute constrained optimization": minimizing the overall sensing error given the dependencies and constraints of multiple attributes of the measured variables is a non-trivial problem to solve; (iii) ``nonuniform task allocation cost": the task allocation cost in social sensing often has a nonuniform distribution which adds additional complexity to the optimized task allocation problem. This paper develops a \textbf{Q}uality-\textbf{C}ost-Aware \textbf{O}nline \textbf{T}ask \textbf{A}llocation (QCO-TA) scheme to address the above challenges using a principled online reinforcement learning framework. We evaluate the QCO-TA scheme through a real-world social sensing application and the results show that our scheme significantly outperforms the state-of-the-art baselines in terms of both sensing accuracy and cost. 
\end{abstract}
 




\end{frontmatter}

	\graphicspath{{pic/}}

	\section{Introduction}

 This paper presents an online reinforcement learning framework to solve the quality-cost-aware task allocation problem in multi-attribute social sensing applications. Social sensing has emerged as a new sensing paradigm in pervasive and mobile computing applications where humans (or devices on their behalf) collectively report measurements about the physical world~\cite{wang2019age,wang2015social}. Examples of social sensing applications include air quality and environment monitoring in smart cities using mobile devices~\cite{wang2015ccs}, malfunctioning urban infrastructures reporting using geotagging~\cite{yu2015mobile}, and damage assessment in disaster response using online social media~\cite{wang2012truth}. In social sensing applications, participants perform sensing tasks at assigned locations to collect different attributes of the measured variables that are of interests to the application~\cite{zhang2018robust}. For example, in an urban air quality sensing application, participants are tasked to measure various air quality attributes (e.g., PM\textsubscript{2.5}, PM\textsubscript{10}, CO\textsubscript{2}) at different locations of the city to estimate the overall air quality and identity potential health risks.  
We refer to this category of applications as \textit{multi-attribute social sensing applications}.

In multi-attribute social sensing applications, there exists a fundamental tradeoff between \textit{data quality} and \textit{sensing (task allocation) cost}~\cite{wang2015ccs,zhang2016incentives}. In particular,  it  is essential to obtain comprehensive and accurate measurements to ensure the desired data quality of the social sensing applications. Such a dedicated data collection process often also encounters a high sensing cost (e.g., more incentives to recruit participants to perform sensing tasks)~\cite{wang2015ccs}. However, the high sensing cost may not always be affordable to the applications with a finite budget~\cite{jaimes2012location}. Therefore, a key challenge for social sensing applications is to find a task allocation strategy (i.e., decide when and where to collect sensing data) that achieves an optimized tradeoff between the data quality and sensing cost. We refer to this problem as the~\emph{quality-cost-aware task allocation} problem. The current solutions to address this problem primarily focus on identifying an \emph{optimal set} of sensing locations (i.e., cells) to collect measurements  to minimize the overall sensing error \cite{wang2015ccs,zhang2018real,tong2016online,zhou2015qoata,hsieh2015inferring, zhang2017expertise,yu2015quality,liu2016cost,liu2018survey}. However, these solutions cannot be directly adapted to solve our task allocation problem due to three challenges that have not been fully addressed: \textit{online task allocation}, \textit{multi-attribute constrained optimization} and \textit{nonuniform task allocation cost}. We elaborate them below.

\textit{Online Task Allocation}. Many social sensing applications are delay sensitive and require timely response to meet the application requirement~\cite{hsieh2015inferring}. For example, during a hurricane, it is crucial for the application to decide when and where the data should be collected to provide real-time situation awareness about the disaster. However, online task allocation in social sensing is challenging due to the large spatial-temporal dynamics of the measured variables and opportunistic nature of social sensing participants~\cite{hsieh2015inferring,zhang2018real}.
This problem becomes more challenging in a multi-attribute sensing scenario where the values of all attributes change simultaneously.
Several task allocation methods have been developed to address similar problems~\cite{wang2015ccs,ahmed2011distance}. However, a few important limitations exist. First, existing models largely ignore the high dynamics of the social sensing applications and allocate the sensing task to cells~\emph{one by one} until the data quality requirement is met~\cite{wang2015ccs}. Second, current solutions do not explicitly consider the correlation between different attributes, thus leading to sub-optimal task allocation solutions.

\textit{Multi-attribute Constrained Optimization}. 
We observe that different sensing attributes often have different spatio-temporal distributions that will affect the task allocation decisions~\cite{zhang2018real}. 
For example, the local optimized task allocation strategy for a particular sensing attribute may not be the global optimized task allocation strategy for all attributes. Furthermore, different sensing attributes may have inherent and complex dependencies. For example, the PM\textsubscript{2.5} and CO\textsubscript{2} are often found to be correlated in a social sensing application that measures the air quality of a city~\cite{zhang2018real}. It is not a trivial job to design a task allocation strategy that can effectively identify the optimal set of sensing cells that can minimize the sensing error across multiple interdependent sensing attributes with diversified distributions.

\begin{figure}[htb!]
	\centering
	\includegraphics[width=0.5\textwidth]{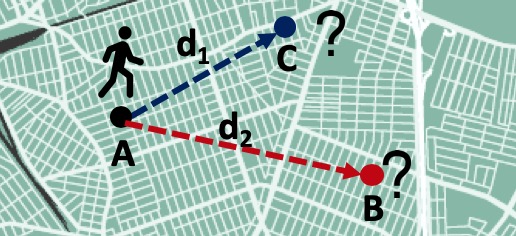}
	\caption{Tradeoff Between Quality and Cost}
	\label{fig:intro_1}
	\vspace{-0.1in}
\end{figure}

\textit{Nonuniform Task Allocation Cost}. The task allocation cost in social sensing is often related to the incentives to motivate a participant to travel from one sensing location to another~\cite{kazemi2012geocrowd}. 
Such task allocation cost often has a non-uniform distribution (e.g., different travel distances will lead to different amount of incentives), which adds additional complexity to the optimized task allocation problem~\cite{zhao2016spatial}. For example, in a social sensing application as shown in Figure~\ref{fig:intro_1}, a participant at location $A$ may be assigned to collect air quality readings at two possible locations: location $B$ and $C$. The sensing measurements collected at B will reduce the overall sensing error more significantly. However, the task allocation cost at $B$ is also higher than $C$ because the travel distance between $A$ and $B$ is larger than the one between $A$ and $C$ (i.e., $d_2 > d_1$). The question is which location we should send participant to perform the sensing task. To answer this question, the task allocation scheme needs to carefully explore the tradeoff between the data quality and the nonuniform task allocation cost.

In this paper, we develop a Quality-Cost-Aware Online Task Allocation (QCO-TA) scheme to address above challenges under a principled online reinforcement learning framework. To address the \textit{online task allocation} challenge, we develop an online learning algorithm that dynamically estimates the priority of each sensing cell for different sensing attributes at each cycle.  
To address the \textit{multi-attribute constrained optimization} challenge, we develop a Bayesian inference scheme that judiciously combines the priority estimations of all sensing attributes into a comprehensive priority score that identifies the cells to effectively reduce the overall sensing error across different sensing attributes.
To address the \textit{nonuniform task allocation cost} challenge, we develop a principled reinforcement learning method to explicitly consider the nonuniform task allocation cost and learns the optimal set of sensing cells for the task allocation.
Finally, we evaluate the QCO-TA scheme on a real-world social sensing dataset: Piemonte Air Quality Dataset. The results show that our scheme significantly outperforms the state-of-the-art baselines in both sensing accuracy and cost.

\yz{We choose the online and reinforcement learning framework to address the quality-cost-aware task allocation problem in multi-attribute social sensing applications for two main reasons. First, the sensing measurements in social sensing applications are often collected in real time~\cite{aggarwal2013social}. The online learning technique is a great fit for such application scenarios because it is capable of dynamically adjusting the task allocation decisions based on the streaming sensing measurements. This is in contrast to the batch based learning techniques which often require a large amount of high-quality training data \emph{a priori} that are not available in our problem setting. Second, our quality-cost-aware task allocation problem aims to find a task allocation strategy that achieves an optimized trade-off between data quality and sensing cost given incomplete sensing measurements (i.e., due to a finite sensing budget). The reinforcement learning technique is a goal-oriented learning technique that fits nicely into our problem. In particular, it provides a data-driven solution that learns to achieve a complex objective (i.e., Equation (2)) given a set of partially available sensing measurements. The reinforcement learning solution sharply contrasts with classical optimization techniques (e.g., linear programming, dynamic programming) which often require a complete set of sensing measurements to learn the optimized trade-off between data quality and sensing cost. 
However, such a complete sensing dataset is often not available due to limited sensing resources and coverage in social sensing applications~\cite{wang2015ccs}.}

A preliminary version of this work was published in the~\cite{zhang2018optimizing}. We refer to the scheme developed in the previous work as the online optimized multi-attribute task allocation (OO-MTA) scheme. The current paper is a significant extension of the previous work in the following aspects. 
First, we extend our previous model by explicitly exploring the optimized tradeoff between the data quality and sensing cost in multi-attribute social sensing applications. In contrast, the OO-MTA only focuses on optimizing data quality and does not take the sensing cost into consideration. 
Second, we develop a novel reinforcement learning algorithm that explicitly addresses the nonuniform task allocation cost challenge identified in this paper. Third, we add a new set of experiments to explicitly evaluate the performance of all compared schemes in terms of both data quality and sensing cost. Fourth, we compare our scheme with more recent task allocation schemes including OO-MTA and demonstrate the performance gains achieved by the QCO-TA scheme compared to all baselines.  Finally, we extend the related work by adding a new discussion on the cost-aware task allocation schemes and the difference between QCO-TA and those schemes (Section~\ref{sec:related}).

	\section{Background and Related Work}
\label{sec:related}

\subsection{Social Sensing}
Social sensing has emerged as a new application paradigm due to the proliferation of portable devices and ubiquitous Internet connectivity~\cite{wang2015social}. A recent survey of social sensing can be found in \cite{wang2019age}. Social sensing has been widely used in environment sensing \cite{hsieh2015inferring}, traffic monitoring \cite{zhang2018risksens}, emergence and disaster response \cite{zhang2017constraint}, social sensor profiling~\cite{zhang2018opinion}, point-of-interest (POI) identification \cite{zhang2017large}, clickbait video detection \cite{shang2019towards}, and abnormal event identification \cite{giridhar2016clarisense+}.  
Quality-cost-aware optimal task allocation in social sensing remains to be an open challenge that has not been fully addressed \cite{wang2015ccs}. This paper addresses the quality-cost-aware task allocation problem in a more challenging scenario where the measured variables have multiple dependent attributes and nonuniform sensing costs. 



\subsection{Task Allocation}

\yz{Task allocation with sparse resources has been well studied in mobile crowdsensing literature~\cite{wang2015ccs,zhang2018real,tong2016online,vance2019towards,zhang2018cooperative,zhou2015qoata,hsieh2015inferring, zhang2017expertise,yu2015quality,liu2016cost,zhang2019integrated,zhang2019heteroedge}. Those techniques can be classified into two main categories based on their primary objectives:  1) \textit{Resource and Cost Reduction}: For example, Wang \textit{et al.} developed a data quality aware task allocation scheme that leverages active learning and Bayesian inference techniques to allocate sensing tasks to a limited number of crowd sensors to reduce the overall sensing cost~\cite{wang2015ccs}. Zhang \textit{et al.} developed a bottom-up task allocation scheme where mobile sensors bid for tasks to minimize energy cost using a game-theoretic approach~\cite{zhang2018real}. Tong \textit{et al.} proposed a two-phased-based online task allocation scheme to reduce the task allocation cost in real-time crowdsourcing systems~\cite{tong2016online}. 2) \textit{Quality-of-Service (QoS) Improvement}: For example, Zhou \textit{et al.} developed a budget-aware task allocation scheme that maximizes the quality of sensing data under the constraints imposed by the physical distance between tasks~\cite{zhou2015qoata}. Hsieh \textit{et al.} developed a greedy task allocation scheme to allocate sensors to cells that would generate minimum entropy to improve the inference accuracy~\cite{hsieh2015inferring}. Zhang \textit{et al.} proposed an expertise-aware task allocation scheme to ensure the quality of the collected data in mobile crowdsourcing systems by inferring the expertise of task participants through truth analysis~\cite{zhang2017expertise}.
There also exist a few solutions that explore both cost and QoS. For example, Yu \textit{et al.} proposed a quality and budget aware task allocation scheme to improve the data quality for spatial crowdsourcing systems given the application specific budget limitations~\cite{yu2015quality}. Liu \textit{et al.} developed an information distribution aware task allocation framework to minimize the task allocation cost of mobile crowdsourcing applications while ensuring the social fairness (e.g., task load balancing) of each participant~\cite{liu2016cost}.}

\yz{
Our work is clearly different from the above solutions in several important aspects. First, we explicitly consider the multi-attribute sensing problem which is more challenging than the single attribute problem addressed in the above literature.
In the multi-attribute sensing problem, different sensing attributes often have different and correlated spatial-temporal distributions that can lead to inconsistent or even conflicting task allocation decisions~\cite{zheng2013u}. For example, the sensing measurements that can significantly improve the estimation accuracy of a specific sensing attribute may not be equally helpful for other sensing attributes.  Moreover, it is not a trivial task to model the complex dependencies between different sensing attributes and understand how such dependencies would affect the global optimized task allocation solution. Second, the goal of our solution is to achieve a complex optimization objective (i.e., jointly optimize the trade-off between the nonuniform task allocation cost and the sensing data quality).
The above objective becomes more challenging when we consider the incomplete sensing measurements introduced by a finite sensing budget in our problem. Such incomplete sensing measurements often provide inadequate evidence to explore the aforementioned trade-off between the data quality and sensing cost in multiple-attribute social sensing applications.
}

\subsection{Online Learning}
\yz{Our work is also related to online learning techniques which have been applied in solving decision making problems in social sensing applications~\cite{feng2010online,zhang2017maintenance,rajan2013crowdcontrol,zhang2018light,xu2017online,zhang2018scalable}. In particular, online learning learns to make sequential decisions to achieve the desired quality-of-service of an application and dynamically adjust the learning process based on the streaming data received at each step. For example, Rajan \emph{et al.} developed an online learning task scheduling scheme that coordinates the real-time execution of crowd tasks through the learning of crowd performance~\cite{rajan2013crowdcontrol}. Feng \emph{et al.} developed an online learning algorithm to detect abnormal behavior patterns in crowds using the online self-organization mapping technique~\cite{feng2010online}. Zhang \textit{et al.} proposed an online learning framework to improve the efficiency of competence-based knowledge compression in machine learning~\cite{zhang2017maintenance}.
Xu \emph{et al.} developed an efficient online learning algorithm for dynamic workload offloading (to the centralized cloud) to minimize the system delay and operation cost~\cite{xu2017online}.  To the best of our knowledge, the QCO-TA scheme is one of the first approaches to leverage online learning techniques to address the quality-cost-aware multi-attribute task allocation problem in social sensing.}

\subsection{Reinforcement Learning}
\yz{Finally, our work also bears some relevance with the reinforcement learning techniques that have been applied in recommendation systems, intelligent transportation systems, computer vision, natural language processing, and control theory~\cite{zhang2017dynamic,xu2018zero,supancic2017tracking,branavan2009reinforcement,abbeel2007application}. In particular, reinforcement learning learns to optimize the desired objective of an application by maximizing the \textit{cumulative reward} received from the environment when exploring the application-specific search space. For instance, Zhang \emph{et al.} developed a scholar collaboration and  recommendation system via competitive multi-agent reinforcement learning~\cite{zhang2017dynamic}.  Xu \emph{et al.} applied a deep reinforcement learning approach in intelligent transportation systems to improve the control robustness for autonomous driving vehicles~\cite{xu2018zero}. Supancic III \emph{et al.} presented a reinforcement learning based decision making framework to continuously track the objects in streaming videos~\cite{supancic2017tracking}. Branavan~\emph{et al.} developed a new reinforcement learning approach to effectively map natural language instructions to executable actions~\cite{branavan2009reinforcement}. Abbeel~\emph{et al.} proposed a reinforcement learning based autonomous helicopter flight control system using differential dynamic programming~\cite{abbeel2007application}. To the best of our knowledge, the QCO-TA scheme is among the first frameworks to leverage reinforcement learning techniques to explore the optimized tradeoff between the data quality and sensing cost in multi-attribute social sensing applications.}



	 \section{Problem Statement}
\label{sec:model}

In this section, we formulate the problem of quality-cost-aware task allocation in multi-attribute social sensing applications. We first define a few terms that will be used in the problem statement.


\begin{definition}
	\textbf{Sensing Cell}: We divide the target area for multi-attribute social sensing task into disjoint cells where each cell represents a subarea of interest. \yz{In particular, we define $S$ to represent the set of sensing cells in the target area, $X$ to be the total number of sensing cells, and $x$ to be the $x^{th}$ sensing cell in the target area.}
	
\end{definition}

\begin{definition}
	\textbf{Sensing Cycle}: A sensing cycle is a period of time where participants  perform one round of the sensing tasks. We define $Y$ to be the total number of sensing cycles, and $y$ to be the $y^{th}$ sensing cycle.
\end{definition}

\begin{definition}
\label{def:sa}
	\textbf{Sensing Attribute}: In social sensing applications, the measured variables often have multiple sensing attributes. We define $A$ to be the total number of sensing attributes, and $a$ to be the $a^{th}$ attribute.
	
\end{definition}
\bigbreak

Consider a social sensing application where the goal is to monitor the air quality index of a city by tasking people to collect sensing measurements at different locations. In this case, a sensing cell is a neighborhood where the sensing values stay relatively stable spatially~\cite{wang2015ccs}. A sensing cycle reflects the frequency of sensing data updates (e.g., hourly, daily). In order to estimate the air quality index, participants  collect a set of sensing attributes (e.g., NO2, CO, PM2.5 and PM10) that are associated with the measured variables (i.e., the air quality index at different cells). 
\bigbreak

\begin{definition}
    \label{def:rs}
	\textbf{Real Sensing Value ($RS$)}: We define the $RS$ matrix to represent the ground-truth sensing value of measured variables.  In particular, $RS^a$ is the ground truth sensing value of attribute $a$, and $RS^a_{x,y}$ is the ground truth sensing value of attribute $a$ in cell $x$ at cycle $y$.
\end{definition}

\begin{definition}
	\textbf{Collected Sensing Value ($CS$)}: We define a $CS$ matrix to represent the collected sensing values of measured variables.  In particular, $CS^a$ is the collected sensing values of attribute $a$, and $CS^a_{x,y}$ is the collected sensing value of attribute $a$ in cell $x$ at cycle $y$.
\end{definition}

\begin{definition}
    \label{def:infer}
	\textbf{Inferred Sensing Value ($IS$)}: We define an $IS$ matrix to represent the inferred sensing values of measured variables from the inference algorithms by leveraging the collected sensing value. In particular, $IS^a$ is the inferred sensing value of attribute $a$, and $IS^a_{x,y}$ is the inferred sensing value of attribute $a$ in cell $x$ at cycle $y$.
\end{definition}

\begin{definition}
    \label{def:sen_error}
    \textbf{Sensing Error ($SE$)}: we define the Sensing Error to be the mean absolute error between the inferred sensing value and the real sensing value. In particular, we have $SE^a_{x,y} = |IS^a_{x,y} - RS^a_{x,y}|$, where $SE^a_{x,y}$ is the inference error of attribute $a$ in cell $x$ at cycle $y$.
\end{definition}

\begin{definition}
\label{def:tp}
\yz{\textbf{Number of Participants ($\mathcal{P}$)}: we define $\mathcal{P}$ to be the total number of participants that can be assigned for task allocation in the application. In social sensing, we observe that the number of available participants is often much smaller than the number of sensing cells $X$ due to the budget and resource limitations~\cite{wang2015ccs}, i.e., $\mathcal{P} << X$. We denote $p$ as the $p^{th}$ participant. In addition, we assume the participants in our multi-attribute social sensing applications to be collaborative, i.e., participants agree to perform all assigned task during the application period as long as compensation is provided.}
\end{definition}


\begin{definition}
\label{def:sen_cost}
\yz{
\textbf{Task Allocation Cost ($\mathcal{C}$)}\footnote{We use the term \emph{task allocation cost} and \emph{sensing cost} interchangeably in the paper.}: we define the task allocation cost to be the compensation to cover the cost of a participant to travel from one sensing cell to another. 
We consider the cost to be proportional to the travel distance of a participant in this paper. In particular, we have
\begin{equation}
    \mathcal{C}^p_y \propto \text{Distance}^p_y
\end{equation}
\noindent where $\mathcal{C}^p_y$ and $\text{Distance}^p_y$ are the task allocation cost and travel distance of participant $p$ at sensing cycle $y$. 
In this paper, we choose the above simplified cost model, however, we observe that the above cost function can be readily extended by considering additional cost/incentive design and mobility modeling~\cite{zhang2015incentives} to accommodate specific requirements of a social sensing application. For example, we can extend the Equation (1) by modeling the diversified response time and transportation expense of different sensing participants, e.g., participants may choose different means of transportation (e.g., walk, bike, drive, bus) to travel between different sensing cells depending on personal preference or availability of different transportation options.}
\end{definition}


The goal of our online quality-cost-aware task allocation in multi-attribute social sensing applications is to make real-time task allocation decisions that optimize the tradeoff between the overall sensing error for all sensing attributes of the measure variables and the task allocation costs over all participants. Formally our problem is defined as: 
\yz{
\begin{equation}
\begin{split}
\label{eq:objective_equ}
&\text{select} \; S_y\;\text{ from } \; S,  \;\forall{1<y<Y_c}\\
&\text{minimize } \sum_{a=1}^{A}(\Gamma_a(\frac{1}{Y_c}\cdot\sum_{y=1}^{Y_c}\frac{1}{X}\cdot\sum_{x=1}^{X}SE^a_{x,y}))\\
&\text{minimize } \frac{1}{Y_c}\cdot\sum_{y=1}^{Y_c}( \frac{1}{\mathcal{P}}\cdot\sum_{p=1}^{\mathcal{P}}\mathcal{C}^p_y)
\\
&\text{where} \enspace |S_y| \leq \mathcal{P} 
\end{split}
\end{equation} 
}
\noindent where $Y_c$ denotes the total number of sensing cycles of interest, and $S_y$ represents the cells that are allocated to the participants for sensing tasks at cycle $y$.
Considering that different sensing attributes have different ranges of values, we define $\Gamma_a$ as the normalization function to normalize the sensing error for attribute $a$.

The above problem is NP-hard since each of its  two objectives (i.e., minimizing sensing error or minimizing sensing cost) 
can be reduced to the Knapsack problem ( i.e., one of the Karp's 21 NP-hard problems)~\footnote{\yz{The detailed proof of NP-hardness of the proposed optimization problem can be found in the Appendix.}}~\cite{pisinger1995minimal}. In this paper, we develop a QCO-TA scheme that judiciously explores the tradeoff between the data quality and sensing cost and identifies an optimized task allocation strategy to jointly optimize the sensing error and cost. The details of the QCO-TA scheme are discussed in the next section.  
	\section{Solution}
\label{sec:solution}

\begin{figure*}[htb!]
\centering
        \includegraphics[width=\linewidth]{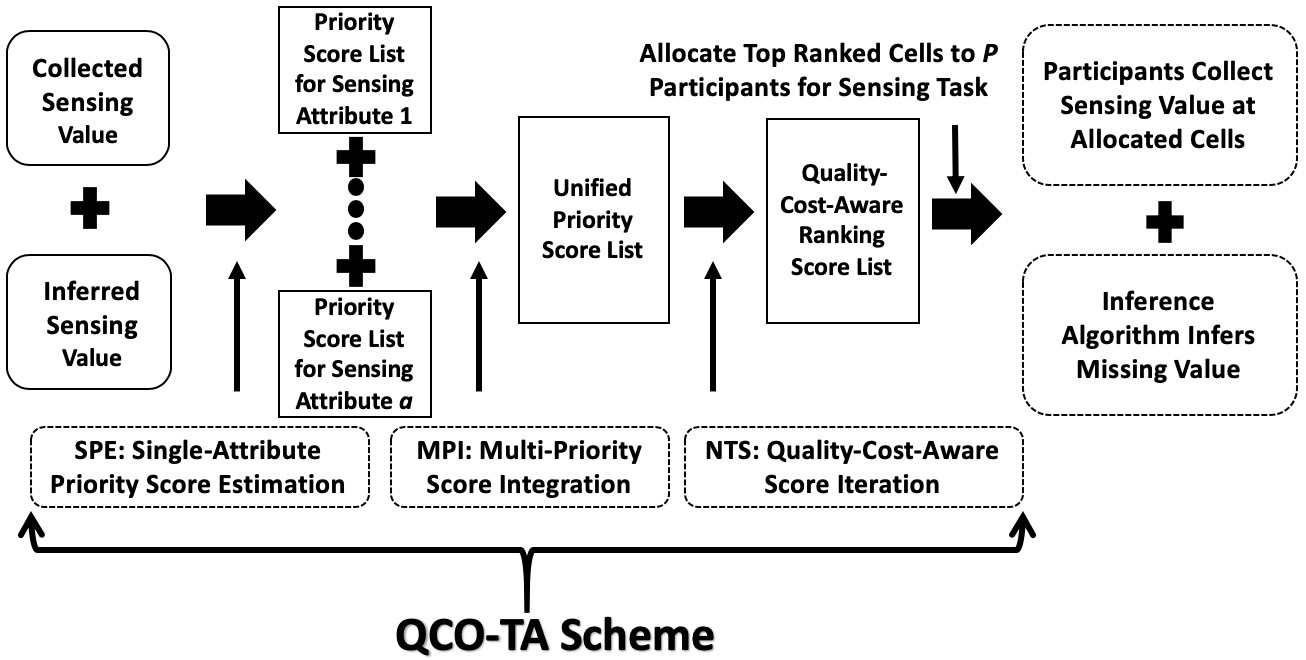}
        \caption{Overview of QCO-TA Scheme}
        \label{overll_arch}
\vspace{-0.2in}
\end{figure*}

In this section, we present the Quality-Cost-Aware Online Task Allocation (QCO-TA) scheme  to address the problem formulated in the previous section using a principled online reinforcement learning framework. The QCO-TA scheme consists of three modules: 1) a Single-Attribute Priority Estimation (SPE) module, 2) a Multi-Attribute Priority Integration (MPI) module, and 3) a Nonuniform-Cost-Aware Task Selection (NTS) module. The overall architecture of the QCO-TA scheme is shown in Figure~\ref{overll_arch}.


\subsection{Single-Attribute Priority Estimation (SPE)}
\label{sec:spe}
In this subsection, we present the single-attribute priority estimation module that addresses the \textit{online task allocation} challenge discussed in the introduction.  We first define a few terms that will be used in this module. 

\begin{definition}

\textbf{Task Priority}: We define the task priority as the order in which the sensing cells are selected for task allocation given a particular sensing attribute \footnote{Since each task is associated with a particular sensing cell, we use task and cell priority interchangeably in the rest of the paper.}. A task with the highest priority will be selected first.
\end{definition}

\begin{definition}
\label{ranking_score}
\textbf{Priority Score}: We further define the \textit{priority score} as a scalar to quantify the task priority defined above.
\end{definition}

In particular, the SPE module estimates the priority score of each cell for a given sensing attribute and dynamically updates the estimations based on the collected sensing values from the previous cycles using an online learning algorithm. 
To compute the task priority of each cell for a given attribute in real time, we need to know which cell's sensing value, if collected, would be the most helpful one to reduce the sensing error. This problem has been proven to be an NP-hard without knowing the real sensing value of cells in advance~\cite{wang2015ccs}. To solve this problem, we develop an efficient approximation algorithm that considers two factors directly related to the sensing error of a cell: 1) \textit{uncertainty}: the estimation confidence of the sensing values in a cell from a given inference algorithm (e.g., KNN, SVR)~\cite{wang2015ccs}; 2) \textit{representativeness}: how accurately the sensing value of the target cell can be used to represent the values of its neighboring cells~\cite{pan2005finding}. In QCO-TA, we use \textit{temporal entropy} and \textit{spatial mutual information} to estimate the uncertainty and representativeness of a sensing cell, respectively.

\textit{Temporal Entropy ($TE$)}: We define temporal entropy to quantify the uncertainty of the inferred sensing value of a sensing cell as follows: 
\begin{equation}
\label{equ:te}
TE_{x,y}^{a} = \Theta_\Lambda(\{IS^a_{x,1},IS^a_{x,2}...IS^a_{x,y}\})
\end{equation}
\noindent where $TE_{x,y}^{a}$ is the temporal entropy of cell $x$ at cycle $y$ for attribute $a$. $IS_{x,y}^a$ is the inferred sensing value of cell $x$ at cycle $y$ for attribute $a$ (defined in Definition~\ref{def:infer}). $\Lambda$ is the distribution (e.g., normal distribution) of the inferred sensing value $IS_{x,y}^a$. $\Theta_\Lambda$ is the function to calculate the differential entropy for the distribution $\Lambda$~\cite{friedman2001elements}. Intuitively, a high temporal entropy of a cell indicates that the inference algorithm is uncertain about its inferred sensing values of that cell and vice versa.


\textit{Spatial Mutual Information (SMI)}: We define the spatial mutual information of a sensing cell to be the aggregated mutual information between the target cell and the rest of cells:
\begin{equation}
\label{equ:smi}
SMI_{x,y}^{a}=\sum_{i=1, i\neq x}^{X}I(\{IS^a_{x,1},IS^a_{x,2}...IS^a_{x,y}\};\{IS^a_{i,1},IS^a_{i,2}...IS^a_{i,y}\})
\end{equation}
\noindent where $SMI_{x,y}^{a}$ represents the spatial mutual information of cell $x$ at cycle $y$ for attribute $a$. $I$ is the function to calculate the mutual information of inferred sensing values between different cells~\cite{ross2014mutual}. In particular, $I(\{IS^a_{x,1},IS^a_{x,2}...IS^a_{x,y}\};\{IS^a_{i,1},IS^a_{i,2}...IS^a_{i,y}\})$ is the mutual information between cell $x$ and $i$. Intuitively, a high spatial mutual information of a cell indicates that the sensing values of the cell, if selected, can be used to reduce the inference error significantly.

We then combine the temporal entropy ($TE$) and spatial mutual information ($SMI$) to compute the \textit{priority score (PS)} that determines the task priority of each sensing cell to be selected for task allocation as follows:
\begin{equation} \label{eq:rank}
PS_{x,y}^{a} = \alpha^y_{TE}\cdot TE_{x,y}^{a}+\alpha^y_{SMI}\cdot SMI_{x,y}^{a}
\end{equation}
\noindent where $PS_{x,y}^{a}$ represents the priority score of cell $x$ at sensing cycle $y$ given sensing attribute $a$. $\alpha^y_{TE}$ and $\alpha^y_{SMI}$ are the weights for temporal entropy and spatial mutual information at cycle $y$, respectively. The values of $\alpha^y_{TE}$ and $\alpha^y_{SMI}$ are tuned based on the requirements of specific applications.

\subsection{Multi-Attributes Priority Integration (MPI)}
\label{sec:mpi}
In this subsection, we describe the Multi-Attribute Priority Integration (MPI) module to address the \textit{multi-attribute constrained optimization} challenge. First, we formally define a comprehensive ranking score as follows.


\begin{definition}
	   \label{ups}
	\textbf{Unified Priority Score ($UPS$)}: We define the $UPS$ to be the weighted summation of the priority score $PS$ of all sensing attributes generated by the SPE module as follows:
	\begin{equation}
        UPS_{x,y}=\sum^{A}_{a=1}w^a_y \cdot PS_{x,y}^{a}
    \end{equation}
    \noindent
    where $UPS_{x,y}$ is the unified priority score for cell $x$ at cycle $y$ and $A$ is the number of sensing attributes. $w^a_y$ is the weight for attribute $a$ in cycle $y$. $PS_{x,y}^{a}$ is the priority score of cell $x$ at cycle $y$ for attribute $a$.
\end{definition}

The key question is how to dynamically compute the weights for all attributes at each cycle so that the aggregated sensing error (defined in Equation~\eqref{eq:objective_equ}) can be minimized by exploiting the dependencies between attributes. To solve this problem, we develop an exponential weighted online learning algorithm that dynamically updates the weights for all attributes based on collected sensing values in real time. In particular, we have the following updating rule for the weight:
	\begin{equation}
	\label{ewa}
        w^a_{y+1}= w^a_{y}\cdot\exp(-\eta \cdot \ell(IS^a_{y},RS^a_{y}))
    \end{equation}
\noindent
where $w^a_y$ and $w^a_{y+1}$ are the weights for attribute $a$ at cycle $y$ and $y+1$, respectively. $\eta$ is the learning rate parameter that directly controls the scale of the weight assigned to each sensing attribute. $\ell(IS^a_{y},RS^a_{y})$ is the loss function that measures the sensing error between the inferred sensing value ($IS^a_{y}$, defined in Definition~\ref{def:infer}) and the real sensing value ($RS^a_{y}$, defined in Definition~\ref{def:rs}) in the current cycle $y$. The intuition of this weight updating function is that it increases the weights of the sensing attributes that contribute less sensing error. 

The challenging part of computing the above weight updating function is how to calculate the loss function since we do not have the real sensing value for all cells due to the budget limitation. To address this problem, we apply Bayesian inference to estimate the loss function as follows: 
	\begin{equation}
        \ell(IS^a_{y},RS^a_{y}) \approx \ell(IS^a_{y},RS^a_{y}|IS^a_{y},CS^a_{y}) = \Phi_{\mathcal{N}_a}(\delta)
    \end{equation}
\noindent
where $\ell(IS^a_{y},RS^a_{y}|IS^a_{y},CS^a_{y})$ is the estimated loss given the collected sensing values $CS^a_{y}$ and inferred sensing values $IS^a_{y}$. $\Phi$ is the inverse of the cumulative distribution function given the distribution $\mathcal{N}_a$. $\mathcal{N}_a$ is the distribution of the mean absolute error ($MAE$)  between $CS_y^a$ and $IS_y^a$ for attribute $a$ in current cycle $y$ (i.e., $|CS_y^a-IS_y^a|$). We assume $\mathcal{N}_a$ follows the normal distribution, which is a common assumption for MAE in social sensing applications~\cite{wang2015ccs}. In addition, $\delta$ is a probability threshold to determine the level of approximation between the loss $\ell(IS^a _{y},RS^a_{y})$ and estimated loss $\ell(IS^a_{y},RS^a_{y}|IS^a_{y},CS^a_{y})$. It is usually set to be higher than 0.95.

	\subsection{ Nonuniform-Cost-Aware Task Allocation (NTS)}
\label{NTS}
In this subsection, we describe the Nonuniform-Cost-Aware Task Allocation (NTS) module to address the \textit{nonuniform task allocation cost} challenge. In particular, the NTS module judiciously integrates the nonuniform task allocation costs with the unified ranking score generated by the MPI module to explore the optimized tradeoff between the data quality and sensing cost through a principled reinforcement learning framework. 
We first define a quality-cost-aware ranking score as follows.
 
\begin{definition}
	\textbf{Quality-Cost-Aware Ranking Score ($QRS$)}: We define the $QRS$ to be the overall ranking score that jointly considers both UPS score (defined in Definition~\ref{ups}) generated by the MPI module and the nonuniform sensing cost (defined in Definition~\ref{def:sen_cost}) as follows:
	\begin{equation}
	    \label{QRS}
        QRS_{x,y}= \Gamma(UPS_{x,y}, \mathcal{C}_{[x',x],y})
    \end{equation}
    \noindent
    \end{definition}
    where $QRS_{x,y}$ is the quality-cost-aware ranking score for cell $x$ at cycle $y$. 
    $UPS_{x,y}$ is the unified priority score of cell $x$ at cycle $y$. $\mathcal{C}_{[x',x],y}$ is the cost for participants to move from current cell $x'$ to cell  $x$ to perform the sensing task at cycle $y$.  
    $\Gamma$ is the mapping function that integrates the UPS score and nonuniform sensing cost. Intuitively, a high QRS value indicates that a cell, if selected, would most likely to reduce the overall sensing errors with the minimal sensing costs, and vice versa.


The key challenge now is how to design an effective mapping function $\Gamma$ to compute the $QRS$ score so that the aggregated sensing error and overall sensing cost can be jointly optimized as indicated in Equation~\eqref{eq:objective_equ}. To solve this problem, we develop a principled reinforcement learning algorithm that iteratively explores the optimal tradeoff between the data quality and the cost for task allocation. We first define a few key terms that will be used in our reinforcement learning framework.

\begin{definition} 
\textbf{State ($\mathcal{S}$)}: We define a state ($\mathcal{S}_x$) as a sensing cell (i.e., $\mathcal{S}_x = x$) that is considered as a candidate for task allocation. Each state carries a state value ($\mathcal{V}$), which indicates the priority of the corresponding cell to be selected for a sensing task allocation. In particular, we define $\mathcal{V}_x^y$ to be the state value for state $\mathcal{S}_x$ at sensing cycle $y$.
\end{definition}
We initiate the state value for each sensing cell with the UPS score generated by MPI at each sensing cycle (i.e., $\mathcal{V}_x^y = UPS_x^y$). The state value will be dynamically updated by our reinforcement learning algorithm to explicitly consider the nonuniform task allocation cost, which will be elaborated in this subsection.
 
\begin{definition} 
\textbf{Action ($\mathcal{A}$)}:
We define an action $\mathcal{A}_{[x',x'']}$ as the move of a participant $p$ who travels from the sensing cell $x'$ to $x''$ to perform the assigned sensing task in two consecutive sensing cycles.
\end{definition}
A participant can take an action to move from current cell $x'$ to a new cells $x''$ or stay at current cell $x'$ (i.e., $x'' = x'$) to perform the sensing task. The goal of our reinforcement learning algorithm is to learn the optimal action for each participant so the aggregated sensing error and overall sensing cost can be jointly optimized.

\begin{definition}  
\textbf{Reward ($\mathcal{R}$)}: We define the reward $\mathcal{R}_{[x',x'']}$ for action $\mathcal{A}_{[x',x'']}$ to be inversely proportional to the task allocation cost (i.e., travel distance $\text{Distance}_{[x',x'']}$ between the two cells) defined in Definition~\ref{def:sen_cost} as:
\begin{equation}
    \mathcal{R}_{[x',x'']} = \gamma\cdot\frac{1}{\text{Distance}_{[x',x'']}}
\end{equation}
\noindent where $\gamma$ is the scaling parameter, the value of $\gamma$ is tuned based on the specific requirements of the applications. The rewards will be used to dynamically update the state values in our  reinforcement learning algorithm.
\end{definition}

Using the above definitions, we leverage a Bellman optimality equation~\cite{bradtke1995reinforcement} to integrate the nonuniform task allocation cost with the unified ranking score generated by MPI as follows:
\begin{equation}
\label{equ:bellman}
\begin{split}
    \mathcal{V}_{x'} = \max_{ \mathcal{A}_{[x',x'']}}\text{E}[\beta\cdot(\mathcal{V}_{x''} )+ R_{[x',x'']}|S = S_{x'},\mathcal{A} = \mathcal{A}_{[x',x'']}]
\end{split}
\end{equation}
\noindent
where $\mathcal{A}_{[x',x'']}$ is the action of moving from the sensing cell $x'$ to $x''$ for a sensing task at sensing cell $x''$. $R_{[x',x'']}$ is the reward for the action. 
$\beta$ is the discount parameter to control the updating rate of the state values, which is usually set to be a small value (i.e., less than 1) to ensure a desired performance of the reinforcement learning algorithm.
The intuition of the above equation is to assign a higher priority score (i.e., state value) to a cell that leads to lower overall sensing error with minimal sensing cost.

Using the Bellman optimality equation, we can learn the optimized state value $V_x^*$ for each sensing cell $x$ by iteratively updating each state value $V_x$ until all state values are converged as follows: 
\begin{equation}
\begin{split}
    abs(\mathcal{V}'_x- \mathcal{V}_x) < \Theta,\forall 1 <x< X\\
    \mathcal{V}^*_x = \mathcal{V}_x,\forall 1 <x< X
\end{split}
\end{equation}
\noindent
where $\mathcal{V}'_x$ is the state value for sensing cell $x$ in the previous iteration, and $\mathcal{V}_x$ is the updated state value for sensing cell $x$ in the current iteration. $\Theta$ is the threshold to stop the learning process, which is usually set to be a small value (e.g., less than 0.1) to ensure the convergence of the learning process and the accuracy of the learned model. We take the learned optimized state value $\mathcal{V}_x^*$ as the overall ranking score $QRS_x^y$ for a sensing cell $x$ at the sensing cycle $y$ as follows: 
\begin{equation}
     QRS_{x,y} = \mathcal{V}^*_x , \forall 1 <x< X
\end{equation}
 

Finally, each participant is allocated to move from the current cell $x'$ to the new cell $x''$  that leads to the highest $ORS_{x'',y}$ value as follows:
\begin{equation}
\label{equ:task_selection}
\begin{split}
    &\argmax_{x''}(ORS_{x'',y}|\mathcal{A}=\mathcal{A}_{[x',x'']}, x''\notin{S_y})\\
    &\text{add } x'' \text{ to } S_y 
\end{split}
\end{equation}
\noindent where $S_y$ is set of sensing cells that has been allocated to the participants in the current sensing cycle. 

	\section{Evaluation}
\label{sec:eval}

In this section, we evaluate the performance of the QCO-TA scheme through a real world social sensing application. We compare the performance of QCO-TA with state-of-the-art task allocation baselines. The evaluation results show that QCO-TA significantly outperforms the baselines in terms of both sensing accuracy and task allocation cost.

\subsection{Datasets}

\begin{figure}[!htb]
\centering
	\subfigure[][Spatial Distribution]{
		\centering
		\includegraphics[width=0.396\textwidth, height=0.324\textwidth]{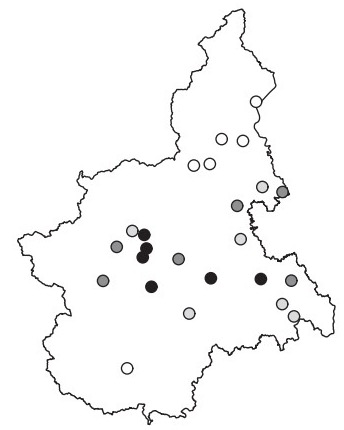}
		\label{fig:sites}
	}
	\subfigure[][Spatial-Temporal Dynamics~\footnotemark]
	{
		\centering
		\includegraphics[width=0.396\textwidth, height=0.324\textwidth]{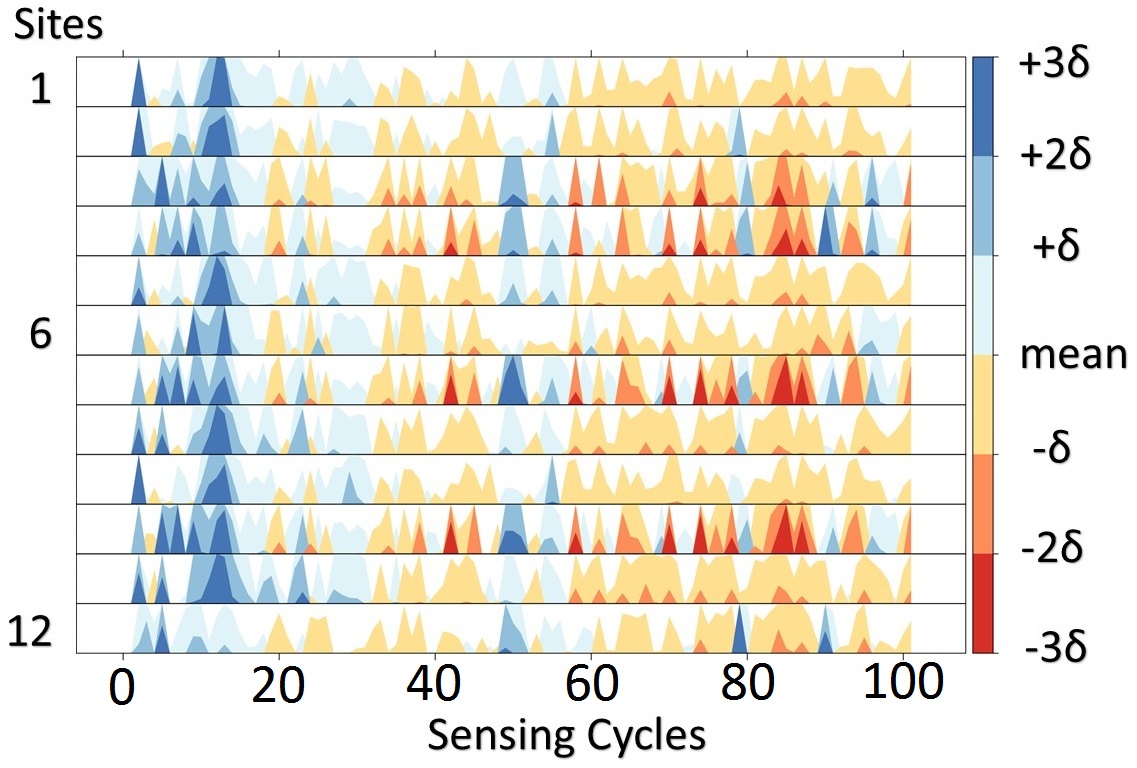}
		\label{fig:sitesDynamic}
	}
	\caption{Piemonte Air Quality Dataset}
	\label{fig:sensors}
	\vspace{-0.1in}
\end{figure}

\textbf{Piemonte Air Quality Dataset:} In our evaluation, we use a social sensing dataset published by Blangiardo \textit{et al.}~\cite{blangiardo2015spatial}~\footnote{https://sites.google.com/a/r-inla.org/stbook/datasets}. 
This dataset consists of daily air quality measurements across 24 locations (i.e., cells) in Piemonte, Italy (as shown in Figure~\ref{fig:sites}). The measurements include the following air quality related attributes: wind speed, temperature, emission rates of primary aerosols, and particulate matter (PM) 10. We choose this dataset because i) it contains multiple sensing attributes of the measured variable (i.e., the air quality); ii) the measurements have large spatial-temporal dynamics (Figure~\ref{fig:sitesDynamic}), which make our problem more challenging to solve. The sensing cycle is set to be one day for this application.

 \footnotetext{The dark colors in each horizontal line indicate larger temporal dynamics and the various color patterns across different lines indicate large spatial variations.}



\subsection{Inference Algorithm}
In the experiment, we select the following inference algorithms to work with the task allocation schemes to estimate the sensing values of the cells that are not selected for sensing in each cycle.

\begin{enumerate}
	\item \textbf{K-Nearest Neighbour (KNN):} KNN estimates the missing value of a cell by averaging the collected sensing values from the \textit{k} nearest cells of the target cell.
	
	
	\item \textbf{Inverse Distance Weighting (IDW):} IDW estimates the missing value of a cell by calculating the weighted average value of the collected sensing values from its \textit{n} closest neighbors, where the weights are proportional to the reciprocal of the spatial distances between the target cell and its neighbors.
	
	
	\item \textbf{Support Vector Regression (SVR):} SVR first establishes a prediction model with the collected sensing values from the selected cells using a support vector machine and then applies the prediction model to infer the missing value of the target cell~\cite{chen2014short}. 
	
\end{enumerate}

\subsection{Baseline algorithms}
We choose several representative task allocation schemes as the baselines. 

\begin{enumerate}
    
    \item \textbf{OO-MTA}: OO-MTA scheme is  a simplified version of QCO-TA scheme that selects the sensing cells solely based on the unified priority score (defined in Definition~\ref{ups}) from our previous work~\cite{zhang2018optimizing}. The OO-MTA scheme does not consider the nonuniform task allocation cost.

    \item \textbf{GPS-TA}: GPS-TA (Greedy Priority Selection Task Allocation) is a greedy task allocation algorithm that selects the $d$ cells with highest priorities for each sensing attribute for task allocation~\cite{hsieh2015inferring}. $d$ is equal to the number of participants divided by the number of attributes. 
    
    \item \textbf{EWA-TA}: EWA-TA (Equal Weighted Aggregation Task Allocation) is a task allocation scheme that generates the overall priority of a cell by calculating the mean of the priorities of the cell across all sensing attributes and then selects the top $\mathcal{P}$ ranked cells for task allocation.
        
    \item \textbf{UNS-TA}: UNS-TA (Uniform Sampling Task Allocation) is a task allocation scheme that uniformly samples \textit{$\mathcal{P}$} cells from all cells for sensing task allocation in each cycle, where each cell has an equal probability to be selected for task allocation~\cite{ho2012online}.     
    
\end{enumerate}

\subsection{Evaluation Metrics}
In our evaluation, we define the following metrics to evaluate the task allocation performance of all compared schemes. 
\begin{itemize}

\item \textbf{Aggregated Sensing Error ($\varepsilon$)}: We define Aggregated Sensing Error ($\varepsilon$) to be the aggregated sensing errors for all the sensing attributes of the measured variable. Specifically, we define:
\begin{equation}
\label{equ:evaluation_metric1}
\varepsilon = \sum_{a=1}^{A}(\Gamma_a(\frac{1}{Y}\cdot \sum_{y=1}^{Y}\frac{1}{X}\cdot \sum_{x=1}^{X}SE^a_{x,y}))
\end{equation}
\noindent 
where $A$ is the number of the sensing attributes, $X$ is the number of sensing cells, and $Y$ is the the number of sensing cycles. $SE^a_{x,y}$ is the sensing error for attribute $a$ of cell $x$ at cycle $y$ as we defined in Definition~\ref{def:sen_error}. $\Gamma_a$ is the normalization function to normalize the sensing error for attribute $a$ as we defined in Equation~\eqref{eq:objective_equ}.

\item \textbf{Average Task Allocation Cost ($\varphi$)}: We define Average Task Allocation Cost ($\varphi$) as follows:
\begin{equation}
\label{equ:evaluation_metric2}
\varphi = \frac{1}{Y}\cdot\sum_{y=1}{Y}( \frac{1}{\mathcal{P}}\cdot\sum_{p=1}^{\mathcal{P}}\mathcal{C}^p_y)
\end{equation}
\noindent where $P$ is the number of participants and $\mathcal{C}^p_y$ is the task allocation cost for participant $p$ at sensing cycle $y$ as we defined in Definition~\ref{def:sen_cost}. In our evaluation, the task allocation cost is measured by the physical distance between the source and destination cells a participant travels. 
\end{itemize}

\subsection{Evaluation Results}
In this subsection, we present the results of our QCO-TA scheme and the compared baselines on the real world social sensing dataset. In the experiment, we evaluate the performance of all compared schemes by varying the number of sensing attributes of the measured variable. In particular, we change the number of of attributes from two to four in our experiment based on the number of attributes available in the dataset (i.e., we have four attributes in total). For a given number of sensing attributes, we evaluate the performance of all compared schemes using the \textit{aggregated sensing error} and \textit{average task allocation cost} metrics defined in Equation~\eqref{equ:evaluation_metric1} and Equation~\eqref{equ:evaluation_metric2}, respectively. In our experiment, we evaluate the performance of all schemes by changing the \textit{number of participants}. Specifically, we vary the number of participants $\mathcal{P}$ from 8 to 14 by considering the number of sensing cells in our dataset (i.e., X=24). 



\subsubsection{Evaluation Results on Data Quality}

\begin{table*}[tbh!]
  \scriptsize  
  \centering
  \caption{Evaluation Results on Aggregated Sensing Error}
  \scalebox{0.85}{
	\begin{tabular}{l|| c | c c c  || c c c  || c c c  }
\toprule
&  & \multicolumn{3}{c}{2 Sensing Attributes} & \multicolumn{3}{c}{3 Sensing Attributes} & \multicolumn{3}{c}{4 Sensing Attributes} \\
\cmidrule(l){2-11}
TA-Scheme & P-Num &KNN&IDW&SVR&KNN&IDW&SVR&KNN&IDW&SVR\\
\midrule
      & $\mathcal{P}$=8 &\textbf{0.153}&\textbf{0.149}&\textbf{0.141}&\textbf{0.184}&\textbf{0.180}&\textbf{0.172}&\textbf{0.232}&\textbf{0.228}&\textbf{0.217}

\\
    \textbf{QCO-TA}&$\mathcal{P}$=10 &\textbf{0.139}&\textbf{0.138}&\textbf{0.134}&\textbf{0.170}&\textbf{0.171}&\textbf{0.166}&\textbf{0.214}&\textbf{0.215}&\textbf{0.211}

\\
     &$\mathcal{P}$=12  &\textbf{0.133}&\textbf{0.131}&\textbf{0.130}&\textbf{0.164}&\textbf{0.163}&\textbf{0.161}&\textbf{0.210}&\textbf{0.209}&\textbf{0.207}
\\
     &$\mathcal{P}$=14  &\textbf{0.129}&\textbf{0.128}&\textbf{0.127}&\textbf{0.160}&\textbf{0.159}&\textbf{0.158}&\textbf{0.207}&\textbf{0.206}&\textbf{0.204}

\\

\cmidrule(l){1-11}
    & $\mathcal{P}$=8 &0.201&0.197&0.175&0.329&0.321&0.263&0.245&0.244&0.231

\\
    OO-MTA & $\mathcal{P}$=10 &0.173&0.170&0.163&0.257&0.252&0.234&0.230&0.230&0.222

\\
    & $\mathcal{P}$=12  &0.163&0.162&0.158&0.230&0.230&0.221&0.226&0.225&0.221

\\
    & $\mathcal{P}$=14  &0.160&0.164&0.159&0.221&0.220&0.214&0.222&0.224&0.219

\\

\cmidrule(l){1-11}
    & $\mathcal{P}$=8 &0.239&0.240&0.221&0.231&0.241&0.226&0.293&0.286&0.275

\\
    GPS-TA & $\mathcal{P}$=10 &0.229&0.232&0.223&0.227&0.228&0.224&0.275&0.275&0.273

\\
    & $\mathcal{P}$=12&0.235&0.242&0.229&0.228&0.232&0.227&0.274&0.277&0.276

\\
    & $\mathcal{P}$=14  &0.238&0.245&0.234&0.232&0.234&0.230&0.276&0.276&0.277

\\

\cmidrule(l){1-11} 
    & $\mathcal{P}$=8 &0.292&0.275&0.245&0.344&0.333&0.289&0.377&0.376&0.334

\\
    EWA-TA & $\mathcal{P}$=10 &0.239&0.234&0.227&0.291&0.290&0.271&0.335&0.335&0.319

\\
    & $\mathcal{P}$=12&0.234&0.226&0.223&0.268&0.264&0.251&0.319&0.318&0.306

\\
    & $\mathcal{P}$=14&0.216&0.210&0.202&0.245&0.240&0.232&0.301&0.299&0.290

\\

\cmidrule(l){1-11}
    & $\mathcal{P}$=8&0.245&0.246&0.223&0.248&0.249&0.229&0.320&0.312&0.288

\\
    UPS-TA & $\mathcal{P}$=10&0.230&0.232&0.223&0.238&0.242&0.232&0.294&0.298&0.289

\\
    & $\mathcal{P}$=12&0.227&0.229&0.224&0.236&0.239&0.234&0.292&0.294&0.289

\\
    & $\mathcal{P}$=14&0.226&0.226&0.222&0.235&0.237&0.233&0.292&0.293&0.290

\\

\midrule
\toprule
    \end{tabular}
    }
\label{tab:task_participants_accuracy}
\end{table*}

\yz{The results on \textit{aggregated sensing error} are shown in Table 1. We observe that the QCO-TA scheme outperforms all of the baselines by achieving the smallest sensing error. The performance gain achieved by the QCO-TA scheme is consistent over different inference algorithms and different numbers of sensing attributes. The performance gain achieved by QCO-TA compared to the best performing baseline on $\mathcal{P}$=8, $\mathcal{P}$=10, $\mathcal{P}$=12, $\mathcal{P}$=14 are 4.8\%, 4.4\%, 3.0\%, 3.1\% under the KNN inference algorithm when the number of sensing attributes is 2. Such performance gains are achieved by two key designs in the proposed QCO-TA scheme. First, the SPE module (described in Section~\ref{sec:spe}) judiciously uses temporal entropy and spatial mutual information to estimate the priority of cells for different sensing attributes.
Two types of sensing cells are often selected for sensing tasks to reduce the overall sensing error: i) the sensing cells with the high uncertainty of their inferred sensing values, and ii) the representative sensing cells whose sensing value can be used to represent the sensing values of their neighboring cells. 
Second, the MPI module (described in Section~\ref{sec:mpi}) unitizes principled exponential weighted online learning to integrate the priority estimations of different sensing attributes. In particular, the MPI module explicitly exploits the dependencies between the attributes and carefully increases the weights of the sensing attributes that contribute less to the overall sensing error. 
Additionally, we observe that the sensing error of all schemes generally decreases when the number of participants increases. This is because a larger number of participants allow the schemes to collect sensing values from more cells, which reduces the errors of the inference algorithms of all compared schemes. These results demonstrate that the QCO-TA scheme can minimize the sensing error of social sensing applications with multiple sensing attributes in comparison with the state-of-the-art baselines.}

\subsubsection{Evaluation Results on Task Allocation Cost}

\begin{table*}[tbh!]
  \scriptsize  
  \centering
  \caption{Evaluation Results on Average Task Allocation Cost (Unit: km)}
  \scalebox{0.85}{
	\begin{tabular}{l|| c | c c c  || c c c  || c c c  }
\toprule
&  & \multicolumn{3}{c}{2 Sensing Attributes} & \multicolumn{3}{c}{3 Sensing Attributes} & \multicolumn{3}{c}{4 Sensing Attributes} \\
\cmidrule(l){2-11}
TA-Scheme & P-Num &KNN&IDW&SVR&KNN&IDW&SVR&KNN&IDW&SVR\\
\midrule
      & $\mathcal{P}$=8 &\textbf{2.872}&\textbf{1.551}&\textbf{5.528}&\textbf{3.043}&\textbf{2.136}&\textbf{6.918}&\textbf{5.696}&\textbf{5.846}&\textbf{6.751}

\\
    \textbf{QCO-TA}&$\mathcal{P}$=10 &\textbf{2.978}&\textbf{5.615}&\textbf{3.136}&\textbf{4.619}&\textbf{4.056}&\textbf{3.787}&\textbf{4.235}&\textbf{5.671}&\textbf{6.008}

\\
     &$\mathcal{P}$=12  &\textbf{3.924}&\textbf{3.789}&\textbf{2.471}&\textbf{3.114}&\textbf{2.779}&\textbf{2.226}&\textbf{3.620}&\textbf{4.259}&\textbf{4.863}
\\
     &$\mathcal{P}$=14  &\textbf{3.281}&\textbf{3.401}&\textbf{2.798}&\textbf{2.563}&\textbf{2.759}&\textbf{3.139}&\textbf{3.116}&\textbf{3.324}&\textbf{4.283}

\\

\cmidrule(l){1-11}
    & $\mathcal{P}$=8&9.338&5.920&11.758&13.810&8.673&7.261&8.310&10.321&8.045

\\
    OO-MTA & $\mathcal{P}$=10&6.234&6.111&6.572&4.916&4.389&8.969&6.585&7.634&6.394

\\
    & $\mathcal{P}$=12&6.557&6.041&5.073&6.303&5.587&6.536&7.378&4.683&5.876

\\
    & $\mathcal{P}$=14&9.560&5.284&4.290&5.571&5.602&5.790&5.619&4.473&4.632

\\

\cmidrule(l){1-11}
    & $\mathcal{P}$=8&16.798&17.436&32.693&29.173&23.686&39.436&20.151&18.421&31.053

\\
    GPS-TA & $\mathcal{P}$=10&13.847&11.883&27.726&16.796&13.283&27.182&20.879&18.253&37.555

\\
    & $\mathcal{P}$=12&10.527&8.250&21.368&14.060&15.727&20.971&17.964&13.922&23.505

\\
    & $\mathcal{P}$=14&7.196&9.840&19.905&15.128&20.066&19.356&13.139&11.631&22.636

\\

\cmidrule(l){1-11} 
    & $\mathcal{P}$=8&11.939&14.584&35.633&14.345&7.202&31.720&15.190&15.925&35.674

\\
    EWA-TA & $\mathcal{P}$=10&9.061&11.515&28.204&10.146&8.822&23.193&10.557&11.685&30.714

\\
    & $\mathcal{P}$=12&6.244&8.726&21.129&5.944&5.577&16.901&8.068&7.868&24.725
\\
    & $\mathcal{P}$=14&7.069&5.664&14.829&5.920&4.226&12.756&5.773&7.736&18.110

\\

\cmidrule(l){1-11}
    & $\mathcal{P}$=8&39.036&36.221&38.862&39.256&36.996&35.553&35.346&35.450&36.655

\\
    UPS-TA & $\mathcal{P}$=10&33.646&32.604&31.180&34.079&31.568&31.678&31.699&35.127&31.713

\\
    & $\mathcal{P}$=12&28.999&28.256&28.083&30.627&30.966&28.356&28.650&28.392&31.266

\\
    & $\mathcal{P}$=14&24.816&23.877&25.082&23.430&25.758&24.251&25.179&25.890&25.301

\\

\midrule
\toprule
    \end{tabular}
    }
\label{tab:task_participants_cost}
\end{table*}

\yz{The results on \textit{average task allocation cost} are shown in Table 2.
We observe that the QCO-TA scheme outperforms all baselines by achieving the lowest task allocation cost. 
The performance gain achieved by QCO-TA compared to the best performing baseline on $\mathcal{P}$=8, $\mathcal{P}$=10, $\mathcal{P}$=12, $\mathcal{P}$=14 are 6.458 km, 3.645 km, 2.633 km, 6.279 km in terms of travel distance under the KNN inference algorithm when the number of sensing attributes is 2.
Similar performance gains are also observed for different inference algorithms and different numbers of sensing attributes. 
Such performance gains of the QCO-TA scheme are achieved by the key design in the principled reinforcement learning framework (NTS module as described in Section~\ref{NTS}), which directly learns to achieve the optimized tradeoff between the data quality and the nonuniform task allocation cost. In particular, the NTS module iteratively updates the proposed Bellman optimality equation (defined in Equation~\ref{equ:bellman}) to assign a higher priority score to the cell that leads to a lower overall sensing error with the minimal sensing cost.
The reinforcement learning process ensures that the QCO-TA scheme always selects the sensing cells which lead to the lowest sensing cost under the premise of sensing quality assurance. 
In summary, the above results demonstrate the capability of QCO-TA to achieve the goal of quality-cost-aware task allocation (defined in Equation~\ref{eq:objective_equ}). We also would like to acknowledge that we only consider the task allocation cost to be the travel distances of participants to perform the sensing tasks in our evaluation. However, in real-world applications, there can be additional factors that affect the task allocation cost (e.g., the time consumed by the participants and the way of commute). We plan to further validate our QCO-TA scheme with a more complex and comprehensive cost model in our future work.
}

\subsubsection{Affordability of QCO-TA scheme}

\begin{figure*}[!thb]

\centering
	\subfigure[][2 Sensing Attributes]{
		\centering
		\includegraphics[width=0.305\textwidth]{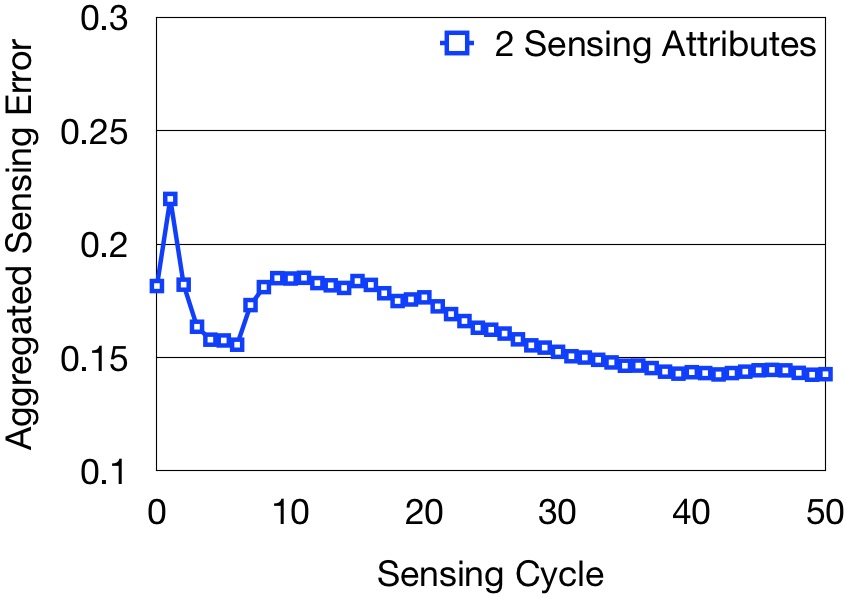}
		\label{fig:2_attr}
	}
	\subfigure[][3 Sensing Attributes]{
		\centering
		\includegraphics[width=0.305\textwidth]{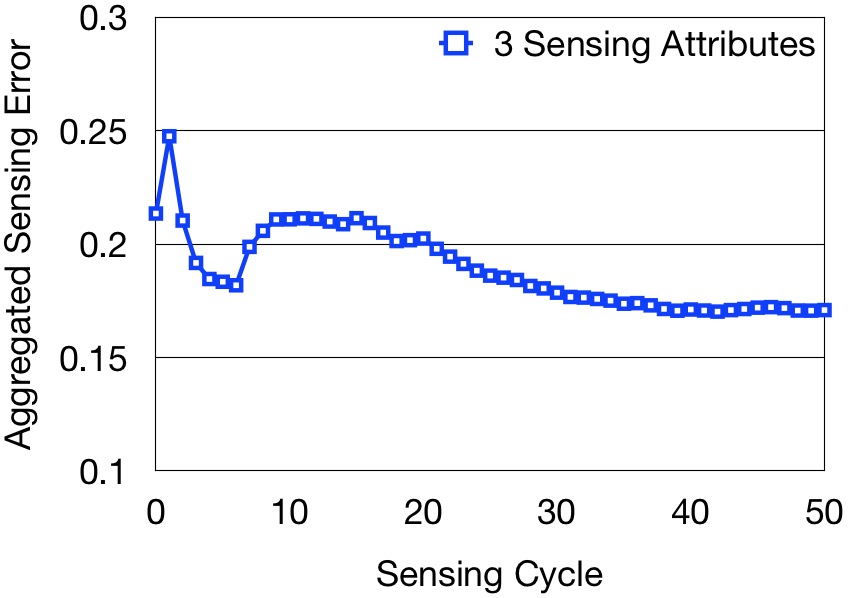}
		\label{fig:3_attr}
    }
        \subfigure[][4 Sensing Attributes]{
		\centering
		\includegraphics[width=0.305\textwidth]{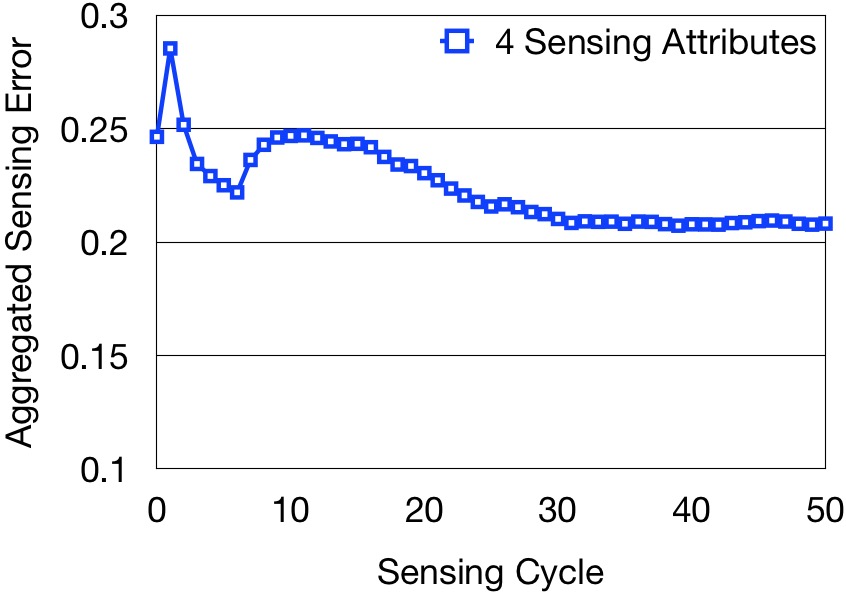}
		\label{fig:4_attr}
	}
	\caption{Affordability of QCO-TA Scheme}
	\label{fig:robust1}

\end{figure*}

\yz{Finally, we study the affordability of the QCO-TA scheme by examining the performance of QCO-TA scheme over multiple sensing cycles. 
The performance of QCO-TA scheme is shown in Figure~\ref{fig:robust1}. We observe that the QCO-TA scheme only requires a small number of sensing cycles at the beginning to quickly learn the optimal task allocation strategy that  optimizes the performance of the system
(in terms of data quality). 
The results demonstrate that our scheme can efficiently learn the optimal task allocation strategy within a limited number of sensing cycles. 
}

	\section{Conclusion}
\label{sec:conclusion}

This paper develops a QCO-TA scheme to solve the quality-cost-aware task allocation problem in multi-attribute social sensing applications. 
In the QCO-TA scheme, we develop a \textit{single-attribute priority estimation} module to estimate the priority score of each cell for a given sensing attribute, a \textit{multi-attribute priority integration} module to integrate the priority scores from all sensing attributes into a unified ranking score for the task allocation, and a \textit{task-cost-aware task allocation} module to explore the optimized tradeoff between the data quality and sensing cost. The evaluation results on a real-world data trace demonstrate that the QCO-TA scheme achieves significant performance gains in terms of both data quality and sensing cost compared to the state-of-the-art baselines in various application scenarios. 

	
\bibliographystyle{elsarticle-num}
\bibliography{ref.bib}
	\yz{\section{Appendix}
\subsection{The proof of NP-hardness of proposed task allocation problem} 
We use the \textit{reduction} technique to prove the quality-cost-aware task allocation problem is NP-hard by reducing it to a well-known NP-hard problem, i.e, the bounded knapsack problem. 
Let us consider a simplified problem of the quality-cost-aware task allocation problem, where the goal is to only optimize the overall sensing error (given the predefined task allocation cost constraint).\newline
\textbf{The proof of NP-hard:} Bounded Knapsack problem is a known NP-complete problem~\cite{pisinger1995minimal}. We do the transformation as follows: 
\begin{enumerate}
    \item For a bounded knapsack problem with a set of $M$ items, each item $m$ comes with a weight of $w^m$ and a value of $v^m$. The goal is to select a subset of at most $N$ items from the $M$ items (i.e., $N<M$) to maximize the overall value of the selected items $\sum^{N}_{i=0}v^i$ while ensuring the overall weight of the selected items is under the knapsack's weight capacity  $\sum^{N}_{i=0}W^i<\Theta$ (where $\Theta$ is the knapsack's weight capacity).
    \item Let us covert the bounded knapsack problem to the simplified task allocation problem as follows: considering a set of $M$ sensing cells, selecting each sensing cell $m$ requires a sensing cost of $w^m$ and reduces the sensing error by $v^m$. The goal is to select a subset of at most $N$ sensing cells from the $M$ sensing cells to maximize the reduced sensing error $\sum^{N}_{i=0}v^i$ while ensuring the overall sensing cost is under the predefined threshold  $\sum^{N}_{i=0}w^i<\Theta$.
    \item By solving the task allocation problem, we can get the answer for the bounded knapsack problem. In particular, if we can allocate $N$ sensing cells to maximize the reduced sensing error while ensuring the overall sensing cost is under the predefined threshold, then we must be able to find $N$ items in the bounded knapsack problem so that the overall value can be maximized while ensuring the overall weight is under the knapsack's weight capacity.
\end{enumerate}
The above reduction process proves that the simplified task allocation problem is NP-hard. In other words, we know that the simplified task allocation problem is at least as hard as all problems in NP, so our proposed quality-cost-aware task allocation problem is also at least as hard as all problems in NP. Therefore, our proposed quality-cost-aware task allocation problem is NP-hard.
}
\end{document}